\documentclass[runningheads]{llncs}
\usepackage[T1]{fontenc}
\usepackage{graphicx}
\usepackage{pgfplots}
\usepackage{lipsum}
\usepackage{subcaption}
\usepackage{algorithm,algpseudocode}
\usepackage{amsmath,amssymb,amsfonts}
\usepackage{comment}
\usepackage{todonotes}

\begin{document}
%
%\title{Contribution Title\thanks{Supported by organization x.}}

% IDEIAS DE TITULOS
%\title{Alleviating catastrophic forgetting with facial expression model strategies}
\title{Alleviating Catastrophic Forgetting in {Facial} Expression Recognition with Emotion-Centered Models}

%\title{Alleviating catastrophic forgetting}
% improving memory retention for facial expression recognition
% alleviating catastrophic forgetting for facial expression recognition
% alleviating catastrophic forgetting with facial-centered models
% alleviating catastrophic forgetting with emotion-centered models
% Alleviating catastrophic forgetting with facial expression model strategies

\titlerunning{Alleviating Catastrophic Forgetting in FER with Emotion-Centered Models}
%\author{Anonymous Authors}
% \orcidID{0000-0002-6171-5777}
\author{Israel A. Laurensi\inst{1} \and
Alceu S. Britto Jr.\inst{1} \and
Jean P. Barddal \inst{1} \and
Alessandro L. Koerich \inst{2}
}

\authorrunning{I. A. Laurensi, A. S. Britto Jr., J. P. Barddal, and A. L. Koerich}

\institute{Graduate Program in Informatics (PPGIa), Pontifícia Universidade Católica do Paraná (PUCPR), Curitiba, PR, Brazil \and
École de Technologie Supérieure (ÉTS), Montréal, Canada}


\maketitle              % typeset the header of the contribution
\begin{abstract}
%\JPBCOMMENT{Notei que a introdução foi alterada para focar na aplicação, mas o abstract ainda não.}
% corrigido
%Catastrophic forgetting, a critical challenge in deep learning, refers to the phenomenon where a neural network loses previously learned information when trained on new tasks. This issue poses significant obstacles in real-world applications that require continuous learning and adaptation to new data. In this paper, we utilize synthetic images generated by Generative Adversarial Networks (GANs) to mitigate catastrophic forgetting in Convolutional Neural Networks (CNNs). Our emotion-centered generative replay (ECgr) method leverages GANs to replicate original training data, thus enabling the CNN to retain knowledge from past tasks while training on new ones. Additionally, we introduce a Quality Assurance (QA) algorithm to filter out poorly generated images by GANs, thereby enhancing the performance of the proposed method. The experimental results on four diverse facial expression datasets demonstrate that incorporating images generated by generative methods enhances training not only on the targeted dataset but also on the source dataset while making the CNN retain previously learned knowledge.

Facial expression recognition is a pivotal component in machine learning, facilitating various applications. However, convolutional neural networks (CNNs) are often plagued by catastrophic forgetting, impeding their adaptability. The proposed method, emotion-centered generative replay (ECgr), tackles this challenge by integrating synthetic images from generative adversarial networks. Moreover, ECgr incorporates a quality assurance algorithm to ensure the fidelity of generated images. This dual approach enables CNNs to retain past knowledge while learning new tasks, enhancing their performance in emotion recognition. The experimental results on four diverse facial expression datasets demonstrate that incorporating images generated by our pseudo-rehearsal meth\-od enhances training on the targeted dataset and the source dataset while making the CNN retain previously learned knowledge.

%\Alceu{Título muito genérico, precisamos citar a aplicação}
%\Alceu{Israel, importante falar no abstract sobre os resultados, impacto. O que observamos?}

\end{abstract}
\section{Introduction}

Emotions are essential in human interaction and comprehension. In such a context, facial expressions play an important role \cite{zavaschi2013fusion}. Thus, facial expression recognition (FER) is the functionality of numerous machine learning applications, including emotion-aware interfaces, personalized recommender systems, and human-robot interaction. One way to identify these emotions in complex systems is through the use of convolutional neural networks (CNNs). These networks have achieved remarkable success in computer vision tasks such as image classification, object detection, and facial expression recognition. However, a significant limitation of CNNs is their susceptibility to catastrophic forgetting. When sequentially trained on different tasks or datasets, CNNs often struggle to retain previously learned information, which leads to degraded performance on previously mastered tasks. This phenomenon impairs the practical application of CNNs in dynamic environments where models must continuously adapt to new data while retaining accuracy in the previous scenarios.

Evaluating the catastrophic forgetting problem in FER - a complex learning scenario - allows us to observe the proposed method's ability to deal with datasets composed of diverse emotional expressions, unlike more straightforward tasks with more limited patterns. Moreover, such an evaluation sheds light on the model's adeptness at maintaining previously learned emotional recognition performance while assimilating the changes of a new domain, showing faces collected with other acquisition protocols, and representing people with different characteristics and cultures.

Catastrophic forgetting arises due to CNN's optimization process, which tends to adjust the model's parameters to fit the current task, often overshadowing previously acquired representations. Numerous approaches have been proposed to mitigate catastrophic forgetting, including regularization techniques, dynamic neural network architecture, and rehearsal-based methods \cite{ewc-related,lwf-related,gr-related,alceu-2018,si-related}. 
Furthermore, several literature reviews have been published in this research field, as well as in continual learning, offering comprehensive insights into the state-of-the-art methodologies, best practices, and emerging trends in mitigating catastrophic forgetting and advancing continual learning algorithms \cite{survey-irina-2022,survey-parisi-2019,survey-vandeven-2022}. While these state-of-the-art methodologies have demonstrated promising results in specific scenarios, they have limitations such as increased computational complexity or limited capacity to effectively retain information from past tasks, especially in facial expression recognition scenarios.

In this paper, we propose a novel approach to overcome the limitations of existing methods and effectively address catastrophic forgetting in CNNs when applied to facial expression recognition. Our approach capitalizes on generative adversarial networks (GANs) capabilities to generate synthetic samples that resemble the original training data. Incorporating these synthetic samples during training enables the CNN to re-learn and retain knowledge from previous tasks, thereby mitigating catastrophic forgetting. To achieve this, we generated synthetic images of each emotion (class) present in the datasets, aiming to better capture the intrinsic characteristics of each facial expression associated with human emotion. We refer to this method as emotion-centered generative replay (ECgr). Moreover, we introduce a quality assurance (QA) algorithm as a crucial component of our approach. The QA algorithm assesses the generated synthetic samples based on the CNN's original classification accuracy. Only the high-quality synthetic samples, which the original CNN can accurately classify, are retained for training. This filtering step ensures that only superior generated samples are utilized, thus augmenting the performance of the proposed method. In addition, we weigh the importance of the synthetic images, considering the CNN output score as an image quality assignment. Such a weight penalizes images that have been assigned a low confidence value by the CNN, which might positively influence the training convergence, as these images may be considered detrimental to the adaptation to the new dataset.

%\Alceu{Destacar também a penalidade do LOSS, pesos diferentes}
% Israel: feito

Our hypothesis centers around the effectiveness of employing a pseudo-re\-hearsal method: H1) the utilization of a pseudo-rehearsal method, particularly our emotion-centered generative replay, offers a potential solution for memory decay in CNNs; H2) the fusion of our emotion-centered generative replay and the proposed QA algorithm offers a promising strategy to counteract memory decay within neural networks; and H3) the combination of emotion-centered generative replay, QA, and a weighted loss function is hypothesized to further strengthen memory retention and performance in neural networks, potentially surpassing the benefits of either ECgr or QA alone. To assess the proposed method's efficiency and validate our hypothesis, we undertook facial expression recognition experiments across various emotion datasets and employed diverse training methodologies.

The contribution of this work is three-fold: i) a new pseudo-rehearsal method focused on the emotions to mitigate catastrophic forgetting when learning facial emotion recognition; ii) a loss function considering a penalization schema for low-quality synthetic images generated in the pseudo-rehearsal strategy; iii) a robust experimental protocol considering well-known FER datasets and a pipeline of experiments to discuss the contributions of the proposed emotion-centered generative replay in mitigating catastrophic forgetting when compared to a regular fine-tuning process or the possibility of joining datasets.

%To evaluate the effectiveness of our approach, we conducted experiments with facial expression recognition tasks using multiple emotion datasets and different training strategies. %The results demonstrate that our method effectively mitigates catastrophic forgetting, surpassing baseline approaches and achieving competitive performance on previous and new tasks.

The remainder of this paper is structured as follows: Section \ref{sec2} reviews related work on catastrophic forgetting and existing methods for its mitigation. Section \ref{sec3} presents the proposed emotion-centered generative replay approach and outlines the architecture of the QA algorithm. Section \ref{sec4} describes the experimental setup and presents the results of our comprehensive evaluations. Section \ref{sec5} discusses the implications of our findings, and Section \ref{sec6} concludes the paper, outlining potential directions for future research.

\section{Related Works}\label{sec2}

%Catastrophic forgetting, an inherent challenge in neural networks, has sparked a wealth of research efforts to mitigate its impact. In this section, we delve into significant algorithms and insights drawn from neuropsychology, all of which strive to tackle the issue of forgetting and enhance memory retention in neural networks.

Catastrophic forgetting has spurred numerous research works to minimize its effects. In this section, we explore prominent algorithms and insights inspired by neuropsychology, all aimed at addressing forgetting and improving memory retention within neural networks.

Learning without forgetting (LWF) stands out by employing knowledge distillation \cite{lwf-related}. This technique transfers distilled knowledge from a model trained on prior tasks to a new model, thereby allowing the assimilation of new information while safeguarding the retention of old information. This intelligent utilization of previous knowledge effectively counteracts the plague of forgetting and amplifies the network's overall performance. Another regularization method, elastic weight consolidation (EWC) \cite{ewc-related}, introduces a nuanced regularization term in the scenario. This term identifies and assigns significance to pivotal network parameters linked to previous tasks, penalizing alterations to these parameters during subsequent training phases. By preserving these key parameters diligently, EWC balances between accommodating novel tasks and upholding the wisdom derived from past experiences. %A practical approach, EWC has proven itself in the battle against catastrophic forgetting.

Synaptic intelligence (SI) \cite{si-related} offers an innovative perspective that stems from evaluating past task performance and assigns weight to synaptic connections based on their influence. The more a synapse contributes, the higher its importance; in contrast, less influential synapses are assigned lower importance. By preserving these critical connections, SI bridges the gap between old and new information, thus mitigating forgetting while embracing novelty.

Deep generative replay (DGR) \cite{gr-related} utilizes generative models to create simulated instances from prior tasks during training. This approach effectively enriches the current task's dataset. The augmented instances, fused with real-time data, offer the network a diverse and comprehensive pool of examples. With past knowledge seamlessly integrated, DGR effectively combats the erosion of previously gained insights, presenting itself as a powerful tool for memory retention.

Beyond these algorithms, insights obtained from neuropsychological research paint a broader picture. Investigations into context-dependent learning have illuminated the crucial role of training and testing contexts in determining network performance. Thus, using contextual cues, algorithms can be designed to exploit the training and testing context better, thereby enhancing memory retention while countering the forgetting phenomenon \cite{survey-parisi-2019}.

The complementary learning system (CLS) hypothesis, which posits dual learning and memory systems within the human brain, further contributes to our understanding. The hippocampal-based system swiftly acquires new knowledge, while the neocortical system excels at long-term storage and retrieval. Incorporating this framework into neural network architectures yields models that deftly balance plasticity and stability \cite{survey-parisi-2019}. By synergizing the strengths of both systems, these models expertly deal with catastrophic forgetting while nurturing the growth of consolidated knowledge.

In light of these contributions, it is crucial to contextualize our work within the broader realm of the current state-of-the-art. The proposed methodology harmonizes the concepts of emotion-centered generative replay and QA. With CNNs as the focal point, our approach aims to prevent catastrophic forgetting in facial expression recognition, a domain where precise emotion identification relies heavily on image quality.

\section{Proposed Method}\label{sec3}

\begin{figure}[!ht]
\includegraphics[width=0.9\linewidth]{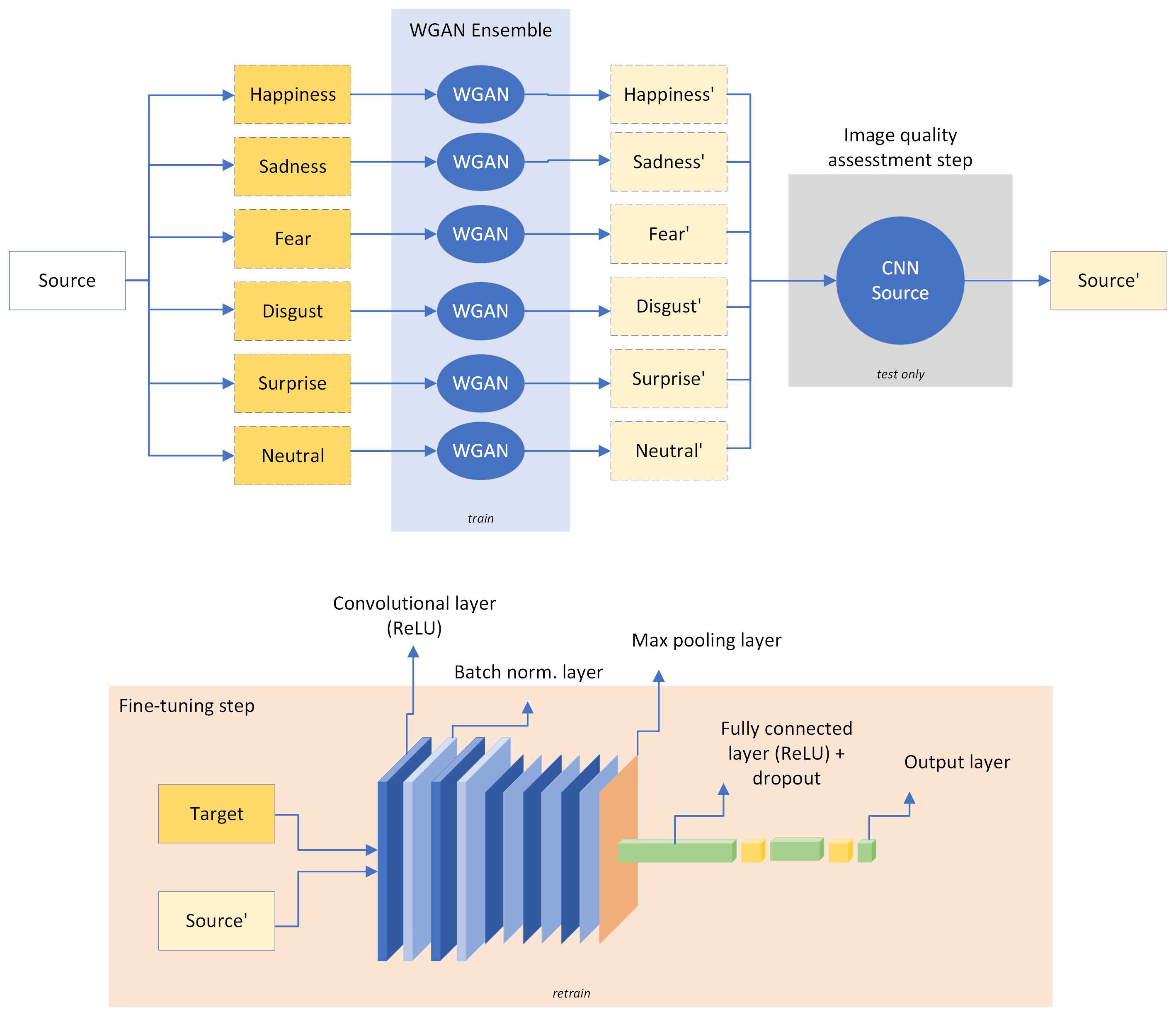}
\centering
\caption{An overview of the proposed method reveals two key components. At the top, the emotion-centered WGAN-GP with CNN QA is depicted. This component involves training a WGAN-GP for each class in the source dataset to generate synthetic data resembling that class. At the bottom, the fine-tuning strategy is illustrated, where our synthetic dataset is replayed alongside the target dataset.}\label{fig:method:genview}
\end{figure}

In this section, we describe the methodology employed in our study to address the challenges of catastrophic forgetting in facial emotion recognition tasks. Our approach combines emotion-centered generative replay using a Wasserstein generative adversarial network with gradient penalty (WGAN-GP) and a QA algorithm. Fig.~\ref{fig:method:genview} presents a general overview of the proposed method.

% TODO
%\Alceu{Aqui vale a pena um  texto bem curto explicando a figura}
% Israel: adicionei no inicio da seção e na legenda da imagem

The use of WGAN-GPs is attributed to the stable learning power of these networks, a factor crucial when dealing with catastrophic forgetting. After all, attempting to address this issue through training and employing a generative method may lead to catastrophic forgetting in the generative networks itself. WGAN-GPs \cite{wgangp-2017} implement a penalty on the gradient norm during training and optimization of the WGAN \cite{wgan-2017}, thereby ensuring more stable training and yielding higher-quality generated images.

%\subsection{Framework and terminology}

%\Alceu{Achei falta daquela figura dando uma visão geral do médoto proposto, lembro que você tem uma}
% Israel: figura adicionada

%To establish a solid foundation for comprehending the intricacies of our approach, we first define several key concepts. These definitions are crucial for understanding the core components of our methodology and how they interact.
%First, let us consider the notion of generic datasets. In our context, datasets are collections of labeled data samples used for training and evaluation. These datasets are organized into classes, each representing distinct categories of data instances. To facilitate the representation of different tasks, we denoted these datasets as $A$, $B$, $C$, and $D$.

%Incorporating WGAN-GPs is another crucial aspect. These networks are a variant of GANs designed stabilize training dynamics. Each WGAN-GP is individually trained to generate synthetic images that accurately represent specific classes.

We have formalized our methodology using algorithmic representations to provide a more concrete understanding of the theoretical concepts presented. In Subsection \ref{method:general_pipeline}, we provide detailed algorithms replicating our approach's offline preparation and training stages. These algorithms encapsulate the step-by-step processes of generating synthetic images and performing continual retraining.

\subsection{Emotion-Centered Generative Replay with WGAN-GP and Quality Assurance}

%\JPBCOMMENT{Minha percepção é que o termo `emotion-centered generative replay' é altamente dependente da aplicação. Nesse sentido, esse termo precisa ser explicado já na introdução pois a aplicação é o que define isso. Quando o termo aparece na introdução pela primeira vez, não fica claro que esse termo está relacionado aos datasets usados na experimentação.}

We initiate by training a set of WGAN-GPs, one for each of the seven emotion classes present in the `source' dataset - fear, anger, happiness, sadness, disgust, surprise, and neutral. Using these trained WGAN-GPs, we generate augmented datasets for each class. These generated images capture the intricate details of respective emotions, diversifying the training data for better generalization.

The WGAN-GP is built by two different networks: discriminator and generator. The discriminator network is crucial in distinguishing between real and synthetic images. It consists of several layers, including convolutional layers with leaky rectified linear united (ReLU) activation functions. These layers help the discriminator extract relevant features from input images. Additionally, dropout layers are applied to prevent overfitting. The discriminator network's output layer produces a single scalar value, representing the likelihood that the input image is a real sample. This network contains approximately 4.3 million trainable parameters, making it a powerful discriminator for the WGAN-GP.

%\JPBCOMMENT{Essa rede é uma proposta sua ou vc tirou de algum lugar? Se for sua, creio que falte detalhes sobre a estrutura da rede (num de camadas, etc). Se for de outro autor, vc precisa adicionar aqui a referência adequada.}
% ISRAEL: Prof., essa rede não é minha, eu já havia utilizado este tipo de rede no mestrado e achei que poderia gerar bons resultados para este contexto, então acabei utilizando elas e foi de fato satisfatório. Adicionei a citação no inicio da seção 3, quando eu menciono as wgangps pela primeira vez.

The generator network is responsible for generating synthetic images. It starts with an input layer that takes random noise as input. This noise is passed through several layers, including dense, batch normalization, and convolutional layers, followed by leaky ReLU activations. These layers progressively upscale and refine the feature maps to generate higher-resolution images. The generator's output shape matches the desired image size. With around 1.6 million total parameters, the generator network creates convincing synthetic images to fool the discriminator.

The generator strives to produce indistinguishable images from real ones, while the discriminator aims to classify them correctly. This adversarial process leads to the training of both networks and ultimately results in the generation of high-quality synthetic data.

We employ our QA algorithm to ensure the quality of the generated images. The QA algorithm filters out low-quality or incorrect images generated by the WGAN-GP, retaining only high-quality images that the original classifier can correctly classify. The QA process is performed using the CNN trained on the source dataset. This network filters the good and bad images from the synthetic datasets. Given an empirically defined threshold, the images correctly classified by the network are used for future retraining, and the misclassified images are discarded. The QA process enhances the reliability of the emotion-centered generative replay, preventing the classifier from being influenced by poor-quality or misleading synthetic images. These images are then integrated into an improved dataset, which merges the synthetic images with the initial source data.

During retraining, the new dataset and the target dataset are employed. This unified dataset facilitates CNN training, where knowledge from the original emotion classes combines with the new target emotions, minimizing forgetting.

% TODO
% \Alceu{Teria uma forma bem enxuta de apresentar a arquitetura da GAN e da CNN?}

\subsection{General Pipeline}\label{method:general_pipeline}

To address the challenge of catastrophic forgetting, our proposed approach involves a two-stage process: offline preparation and a training phase.

\subsubsection{Offline preparation stage}\label{method:offline_stage}

%\JPBCOMMENT{Revisar texto e ajustar a escrita pois o algoritmo é que faz as coisas e não nós (we)}
% corrigido

Initialization occurs as depicted in Algorithm \ref{alg:offline_stage}, where a set of datasets represented by $T$ is defined, encompassing datasets $A, B, C$, and $D$. Each dataset $d_t$ within $T$ is traversed through an iterative process. For each specific dataset $d_t$, a classifier, denoted as $C_{d_t}$, is trained using that particular dataset.

% algoritmo offline stage
\begin{algorithm}[!htb]
    \caption{Offline stage}\label{alg:offline_stage}
    \begin{algorithmic}[1]
        \State $T \gets A, B, C, D$
        \State $T' \gets \emptyset$
        \For{each dataset $d_t$ in $T$}
            \State $G_{d_t} \gets \emptyset$
            \State Train classifier $C_{d_t}$ on dataset $d_t$
            \For{each class $c$ in dataset $d_t$}
                \State Train $\texttt{WGANGP}_c$ on class $c$
                \State Add $\texttt{WGANGP}_c$ to ensemble $G_{d_t}$
                \State Generate $\texttt{SI}_c$ using $\texttt{WGANGP}_c$
                \State Pass $\texttt{SI}_c$ through $C_{d_t}$ to generate dataset ${d_t}_{c}^{qa}$
                \State Add ${d_t}_{c}^{qa}$ to dataset $d_t'$
            \EndFor
        \State Add $d_t'$ to $T'$
        \EndFor

        \State \textbf{return} collection of synthetic datasets $T'$
    \end{algorithmic}
\end{algorithm}

%\JPBCOMMENT{No algoritmo acima, use $\texttt{WGANGP}_c$ para não deixar ambíguo para o leitor. O mesmo vale para $SI$.}
% corrigido

%\JPBCOMMENT{$T'$ parece não ser inicializado ($T' \leftarrow \emptyset$, por exemplo)}
% corrigido

%\JPBCOMMENT{os símbolos matemáticos abaixo (no parágrafo) também precisam de ajustes.}
% corrigido

\vspace{-10pt}

We then iterate over each class $c$ in dataset $d_t$. In this context, we train a WGAN-GP for each class, denoted $\texttt{WGANGP}_c$, and subsequently combine them to form an ensemble, denoted as $G_{d_t}$. Through these $\texttt{WGANGP}_c$, we generate synthetic images ($\texttt{SI}_c$) mirroring the characteristics of each class. Continuing the process, we feed these synthetic images through the classifier $C_{d_t}$, resulting in a new dataset, ${d_t}_{c}^{qa}$, consisting only of images correctly classified by $C_{d_t}$. These refined synthetic images are combined into a new dataset $d_t'$. This procedure is executed for each dataset $d_t$, and all resulting datasets $d_t'$ are unified into a collection labeled as $T'$, encapsulating the sets of synthetic datasets corresponding to each original dataset $d_t$ in $T$.

\subsubsection{Continual learning stage}\label{method:training_stage}

Our approach began with individual training for ECgr before merging ECgr and QA. For a comparative evaluation, we utilize joint training and fine-tuning methods. Joint training simultaneously incorporates the source and target data while fine-tuning adapts the CNN to new data, training only the fully connected layers.

% algoritmo continual learning
\begin{algorithm}[!htb]
    \caption{Continual learning stage}\label{alg:training_stage}
    \begin{algorithmic}[1]
        \State $T \gets B, C, D$
        \State $C_T \gets \emptyset$
        \For{each dataset $d_t$, $d_t'$ in $T$, $T'$}
            \State $d_{t}^{u} \gets d_t + d_t'$
            \State Train classifier $C_{d_{t}^{u}}$ on unified dataset $d_{t}^{u}$
            %\State Apply $EWC(C_{t_s}^{d})$ regularization
            \State Add trained $C_{d_{t}^{u}}$ to $C_T$
        \EndFor

        \State \textbf{return} ensemble $C_T$
    \end{algorithmic}
\end{algorithm}

As shown in Algorithm \ref{alg:training_stage}, we define a set of subsequent datasets, indicated by $T$, which comprises datasets B, C, and D. Then, we iterate over each combination of the original dataset and subsequent dataset, referred to as $d_t$ and $d_t'$ respectively, from set $T$ and its counterpart $T'$. For each dataset combination, we create a unified dataset, $d_{t}^{u}$, by merging $d_t$ and $d_t'$. Subsequently, we train a classifier, $C_{d_{t}^{u}}$ on the unified dataset $d_{t}^{u}$. Finally, we accumulate the trained classifiers $C_{d_{t}^{u}}$ for each combination, and this ensemble is returned as $C_T$, representing a comprehensive collection of classifiers suited for different datasets and training strategies.

The CNN used in the experiments comprises several layers, starting with 2D convolutional layers with 64 filters each. Batch normalization stabilizes and accelerates training, followed by additional convolutional layers with varying filter sizes. Max-pooling layers are introduced to downsample the feature maps, reducing the spatial dimensions and focusing on the most salient features. The network continues with more convolutional layers and batch normalization to extract higher-level representations from the input data. Eventually, the feature maps are flattened to form a one-dimensional vector, which is then passed through fully connected layers. Dropout layers are employed to prevent overfitting, enhancing the model's generalization. The final layer is dense with a softmax activation function, producing output probabilities for different classes. In total, this CNN architecture contains approximately 19.3 million parameters.

The training aims to optimize the Eq.~\eqref{eq:loss-weighted}, where a weight $w$ is applied to each prediction. This weight is determined by the CNN's confidence percentage when predicting for all $y_{pred}$.

% equação do loss
\begin{equation}
\label{eq:loss-weighted}
%\begin{align*}
\mathcal{L}_i(\mathbf{y}_{\text{true}}^{(i)}, \mathbf{y}_{\text{pred}}^{(i)}) = -\sum_{j=1}^C w_j y_{\text{true j}}^{(i)} \log(y_{\text{pred j}}^{(i)})
%\end{align*}
\end{equation}

In summary, our general pipeline encompasses an offline preparation phase involving training WGAN-GPs and QA-based synthetic image generation. In the training stage, augmented and original datasets are combined, and the continual retraining approach adapts the classifier to multiple datasets while incorporating different strategies.

\section{Experiments}\label{sec4}

%\subsection{Datasets}

To evaluate the performance of our methodology, we utilize several datasets that contain human facial images displaying various emotions. The datasets considered in our study include TFEID, MUG, CK+, and JAFFE. These datasets provide a diverse set of emotional contexts, allowing us to assess our approach's robustness and generalization capabilities. All datasets have the following classes: fear, anger, happiness, sadness, disgust, surprise, and neutral.

%\JPBCOMMENT{As figuras abaixo não são citadas no texto e precisam ser.}
% figuras removidas para ocupar menos espaço

The multimodal understanding group (MUG) \cite{mug-dataset} dataset consists of approximately 1462 facial images, each annotated with the corresponding facial expression labels. The Japanese female facial expression (JAFFE) \cite{jaffe-dataset} dataset, despite its relatively small size, containing approximately 213 facial images, is valuable for evaluating and comparing facial expression recognition models. The Taiwanese facial expression image database (TFEID) \cite{tfeid-dataset} provides a suitable testbed for evaluating emotion recognition algorithms, with 1128 samples. Lastly, the extended Cohn-Kanade dataset (CK+) \cite{ckold-dataset} dataset is commonly used for facial expression recognition research. It includes a substantial number of facial images, compiled into 123 videos of different subjects, totaling approximately 593 videos, with 327 labeled videos covering various emotional expressions. 

\subsection{Results}

This section offers an in-depth analysis of the outcomes achieved by employing different retraining strategies, each suited to minimize memory degradation and maximize knowledge retention. %Through a systematic exploration of accuracy metrics across various datasets and scenarios, we discuss valuable insights into the effectiveness of the strategies in adapting to new tasks while preserving knowledge from previous ones.

\vspace{-20pt}

\subsubsection{On Quality of Synthetic Images}\label{results:wgan}

\begin{figure}[!ht]
\includegraphics[width=0.80\linewidth]{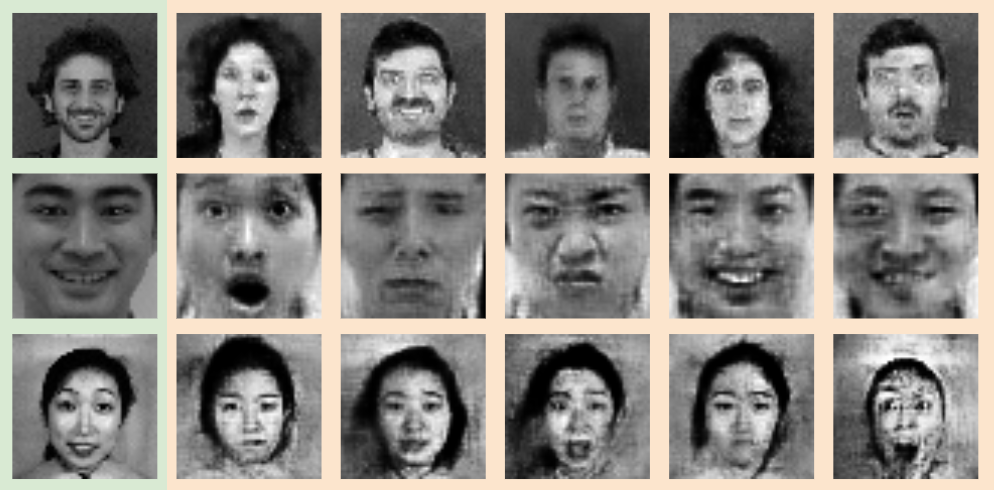}
\centering
\caption{Result samples on different classes from the MUG, JAFFE, and TFEID's synthetic dataset generated by the WGAN-GP. The first column (in green) displays the original samples from the MUG, TFEID, and JAFFE datasets from top to bottom, respectively. In contrast, the second-to-last column (in orange) features the corresponding synthetic images for each dataset.}\label{results:synthetic-images}
\vspace{-10pt}
\end{figure}

%\vspace{-15pt}

In addition to quantitative assessments, we also conducted a qualitative analysis of the synthetic images generated through our approach. In this subsection, we present a comprehensive discussion of the results, including a thorough examination of the qualitative aspects of the synthetic data. This analysis provides valuable insights into the visual fidelity and relevance of the generated images. As shown in Fig.~\ref{results:synthetic-images}, the left side features an image from the original dataset as a reference for the dataset's inherent visual characteristics. On the right side, seven columns display synthetic images generated for each class within the dataset. These columns showcase the diversity and fidelity of the synthetic samples produced by our ECgr approach.

\begin{figure}[!ht]
\vspace{-10pt}
\includegraphics[width=0.80\linewidth]{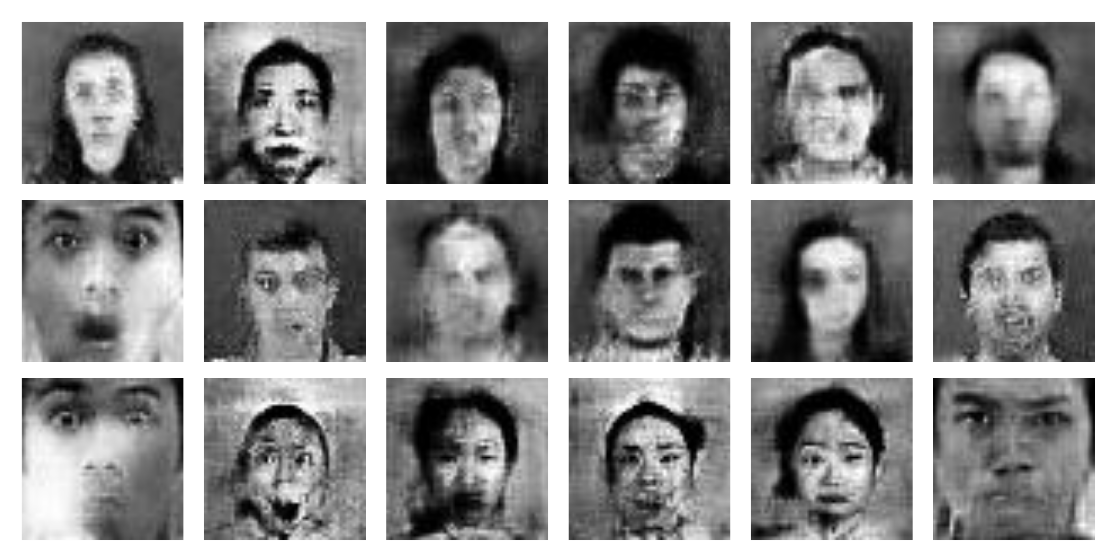}
\centering
\caption{Some rejected samples identified by the QA algorithm from the synthetic datasets of MUG, JAFFE, and TFEID.}\label{results:rejeitadas}
\vspace{-20pt}
\end{figure}

The results obtained using QA filtering can also be evaluated qualitatively. In Fig.~\ref{results:rejeitadas}, we can see some examples of images that were rejected during the QA process. These images are of low quality and do not demonstrate emotion; therefore, the CNN could not classify them correctly.

\subsection{On Continual Learning}

In this section, we will discuss the main results observed from the tests conducted with facial expression datasets, utilizing the combination of different methods outlined in this study.

Initially, a CNN was trained on the MUG dataset, and subsequently, we employed this CNN to adapt it to other datasets continuously. All trainings were conducted with 20 replications. In methods involving image generation, each experiment replication utilizes images explicitly generated for that retraining; in other words, synthetic datasets are not identical across different replications of CNN adaptation.

In Tables \ref{results:table_mug_to_jaffe}, \ref{results:table_mugjaffe_to_tfeid}, and \ref{results:table_mugjaffetfeid_to_ckold}, the columns baseline, joint, and fine-tune represent, respectively: testing datasets with the CNN trained on the source dataset; adapting the CNN trained on the source plus target dataset; adapting the CNN trained on the source dataset using only the target dataset. Additionally, the ECgr and QA methods were separately evaluated and combined to ascertain the impact of using synthetic image filtering in continuous training. Furthermore, this scenario assessed whether using weights on synthetic images has any effect compared to training without this technique.

%Table 1
\begin{table}[!htb]
\centering
\caption{Results on MUG's model fine-tuned to JAFFE dataset in terms of ECgr, QA, weighted QA, and the combination of ECgr with QA and wQA, alongside with fine-tune, joint, and current for a direct comparison.}\label{results:table_mug_to_jaffe}
\begin{tabular}{c|c|c|c|c|c|c}
\hline
\multicolumn{1}{l|}{\textbf{}} & \textbf{Current}   & \textbf{Joining}      & \textbf{Fine}  &
\multicolumn{3}{c}{\textbf{Proposed}} \\
\cline{5-7}
\multicolumn{1}{l|}{\textbf{}} & \textbf{model}   & \textbf{datasets}      & \textbf{Tuning}  & \textbf{ECgr}        & \textbf{ECgr+QA}     & \textbf{ECgr+wQA}    \\ 
\hline
\multicolumn{7}{l}{\textbf{Source dataset}} \\ 
\hline
MUG                   & 0.98$\pm$0.00 & 1.00$\pm$0.00 & 0.75$\pm$0.03 & 0.88$\pm$0.04 & 0.93$\pm$0.02 & 0.94$\pm$0.03 \\ \hline
\multicolumn{7}{l}{\textbf{Target dataset}} \\
\hline
JAFFE                 & 0.28$\pm$0.00 & 0.74$\pm$0.06 & 0.77$\pm$0.03 & 0.78$\pm$0.03 & 0.78$\pm$0.05 & 0.79$\pm$0.04 \\ \hline
Mean                  & 0.63       & 0.87       & 0.76       & 0.83       & 0.85       & 0.86 \\ \hline      
\end{tabular}
\vspace{-10pt}
\end{table}

In Table \ref{results:table_mug_to_jaffe}, it is possible to observe the results when adapting the CNN trained on the MUG dataset (source) to the JAFFE dataset (target). Taking into account the baseline, joint, and fine-tune methods, we can assume that the upper limit is the joint method, which represents the ideal case where all datasets are available for training, and the lower limit is the fine-tune method, in which the source dataset is no longer available. It can be concluded that the combined method ECgr+wQA yielded the best results in this initial adaptation involving only one dataset, with a result very close to joint training. In Tables \ref{results:table_mugjaffe_to_tfeid} and \ref{results:table_mugjaffetfeid_to_ckold}, we will notice a change in this scenario, as more datasets are introduced in continuous training, the ECgr+QA method tends to outperform. Regarding the result obtained in the retraining for the JAFFE dataset, it is possible to justify this outcome, where the combined method came close to the joint method, as the adaptation can still be considered trivial since only one dataset is being adapted. Thus, the complexity for the CNN to assimilate synthetic images remains low.

% TODO
%\Israel{TODO: Deixar mais claro no texto a comparação entre as means observadas nas tabelas}

%Table 2
\begin{table}[!b]
\vspace{-20pt}
\centering
\caption{Results on MUG plus JAFFE's model fine-tuned to TFEID dataset in terms of ECgr, QA, weighted QA, and the combination of ECgr with QA and wQA, alongside with fine-tune, joint and current for a direct comparison.}\label{results:table_mugjaffe_to_tfeid}
\begin{tabular}{c|c|c|c|c|c|c}
\hline
\multicolumn{1}{l|}{\textbf{}} & \textbf{Current}   & \textbf{Joining}      & \textbf{Fine}  &
\multicolumn{3}{c}{\textbf{Proposed}} \\
\cline{5-7}
\multicolumn{1}{l|}{\textbf{}} & \textbf{model}   & \textbf{datasets}      & \textbf{Tuning}  & \textbf{ECgr}        & \textbf{ECgr+QA}     & \textbf{ECgr+wQA}    \\ \hline
\multicolumn{7}{l}{\textbf{Source datasets}} \\ 
\hline
MUG                  & 0.75$\pm$0.03 & 1.00$\pm$0.00 & 0.71$\pm$0.01 & 0.84$\pm$0.06 & 0.87$\pm$0.04 & 0.78$\pm$0.04 \\
JAFFE                & 0.77$\pm$0.03 & 0.94$\pm$0.03 & 0.76$\pm$0.02 & 0.64$\pm$0.07 & 0.62$\pm$0.07 & 0.69$\pm$0.04 \\ \hline
Mean                  & 0.75       & 0.97       & 0.73       & 0.74       & 0.74       & 0.73\\ \hline

\multicolumn{7}{l}{\textbf{Target dataset}} \\ 
\hline
TFEID                 & 0.22$\pm$0.00 & 0.79$\pm$0.05 & 0.78$\pm$0.03 & 0.83$\pm$0.04 & 0.84$\pm$0.04 & 0.87$\pm$0.04 \\ \hline
Updated mean          & 0.58       & 0.91       & 0.75       & 0.77       & 0.78       & 0.78 \\ \hline      
\end{tabular}
\vspace{-20pt}
\end{table}

Table \ref{results:table_mugjaffe_to_tfeid} shows that the best result lies between ECgr+QA and ECgr+wQA. Interestingly, in all results, the generative method - combined or not - performed equally or better on the target dataset when compared to joint training. This reveals that synthetic images not only aid the CNN in recalling something it has already seen but also assist in training for new data, reinforcing knowledge when adapting to the same context, in this case, facial expression recognition.

\begin{table}[!htb]
\centering
\caption{Results on MUG plus JAFFE plus TFEID's model fine-tuned to CK+ dataset in terms of ECgr, QA, weighted QA, and the combination of ECgr with QA and wQA, alongside with fine-tune, joint and current for a direct comparison.}\label{results:table_mugjaffetfeid_to_ckold}
\begin{tabular}{c|c|c|c|c|c|c}
\hline
\multicolumn{1}{l|}{\textbf{}} & \textbf{Current}   & \textbf{Joining}      & \textbf{Fine}  &
\multicolumn{3}{c}{\textbf{Proposed}} \\
\cline{5-7}
\multicolumn{1}{l|}{\textbf{}} & \textbf{model}   & \textbf{datasets}      & \textbf{Tuning}  & \textbf{ECgr}        & \textbf{ECgr+QA}     & \textbf{ECgr+wQA}    \\ \hline
\multicolumn{7}{l}{\textbf{Source datasets}} \\ 
\hline
MUG                   & 0.71$\pm$0.01 & 1.00$\pm$0.00 & 0.63$\pm$0.05 & 0.85$\pm$0.04 & 0.87$\pm$0.03 & 0.73$\pm$0.04 \\
JAFFE                 & 0.76$\pm$0.02 & 0.99$\pm$0.01 & 0.57$\pm$0.08 & 0.61$\pm$0.05 & 0.55$\pm$0.04 & 0.59$\pm$0.05 \\
TFEID                 & 0.78$\pm$0.03 & 1.00$\pm$0.00 & 0.49$\pm$0.06 & 0.62$\pm$0.09 & 0.76$\pm$0.09 & 0.70$\pm$0.07 \\ \hline
Mean                  & 0.73       & 0.95       & 0.56       & 0.69       & 0.72       & 0.67\\ \hline
\multicolumn{7}{l}{\textbf{Target dataset}} \\ 
\hline
CK+                 & 0.53$\pm$0.00 & 0.81$\pm$0.03 & 0.79$\pm$0.03 & 0.83$\pm$0.03 & 0.82$\pm$0.03 & 0.81$\pm$0.02 \\ \hline
Updated mean                  & 0.68       & 0.99       & 0.62       & 0.72       & 0.75       & 0.71 \\ \hline
\end{tabular}
\vspace{-10pt}
\end{table}

In Table \ref{results:table_mugjaffetfeid_to_ckold}, we can observe a behavior similar to that observed when adapting to TFEID, where the best result lies between the ECgr+QA and ECgr+wQA methods. However, at this point, it becomes more apparent that using a weight for synthetic images brings an intrinsic problem to the training of the CNN being used for the filtering method. This CNN can carry certain behaviors into subsequent training steps, where errors from certain classes may compromise the entire training when using the confidence percentage.

% TODO
%\Alceu{Israel, podemos afirmar que a JAFFE é a que mais sofre perda durante o aprendizado incremental? Se sim, isto estaria relacionado à característica desta base - pouca diversidade (apenas mulheres, japonesas)}

\begin{figure}[!ht]
\vspace{-10pt}
    \centering
    \begin{tikzpicture}[scale=0.8]
    \definecolor{ptblue}{RGB}{0,68,136}
    \definecolor{ptred}{RGB}{187,85,102}
    \definecolor{ptyellow}{RGB}{221,170,51}
    \definecolor{ptgreen}{RGB}{34,136,51}
        \begin{axis}[
            clip=false,
            ylabel={Accuracy},
            ymin=0.6,
            %legend pos=outer north east,
            xtick=data,
            xticklabels={Baseline, JAFFE, TFEID, CK+},
            ytick={0.6, 0.65, 0.7, 0.75, 0.8, 0.85, 0.9, 0.95, 1},
            axis lines=left,
            xmajorgrids=true,
            ymajorgrids=true,
            grid style=dashed,
            ]
            \addplot[color=ptblue,mark=*] coordinates {
                (1, 0.9800)
                (2, 1.0000)
                (3, 1.0000)
                (4, 1.0000)
            }node[right, pos=1.03]{Joining datasets};
            %\addlegendentry{Joint}
            
            \addplot[color=black,mark=x] coordinates {
                (1, 0.9800)
                (2, 0.7500)
                (3, 0.7100)
                (4, 0.6300)
            }node[right, pos=1.03]{Fine-tuning};
            %\addlegendentry{Fine-tune}
            
            \addplot[color=ptred,mark=triangle*] coordinates {
                (1, 0.9800)
                (2, 0.8800)
                (3, 0.8400)
                (4, 0.8500)
            }node[right, pos=1.03]{ECgr};;
            %\addlegendentry{ECgr}
            
            \addplot[color=ptgreen,mark=square*] coordinates {
                (1, 0.9800)
                (2, 0.9300)
                (3, 0.8700)
                (4, 0.8700)
            }node[right, pos=1.03]{ECgr+QA};;
            %\addlegendentry{ECgr+QA}
            
            \addplot[color=ptyellow,mark=diamond*] coordinates {
                (1, 0.9800)
                (2, 0.9400)
                (3, 0.7800)
                (4, 0.7300)
            }node[right, pos=1.03]{ECgr+wQA};;
            %\addlegendentry{ECgr+wQA}
        \end{axis}
    \end{tikzpicture}
    \caption{Accuracy results on the MUG dataset, showcasing the continuous adaptation of a trained CNN across JAFFE, TFEID, and CK+ datasets relative to the baseline accuracy.}
    \label{results:accuracy_mug}
    \vspace{-10pt}
\end{figure}
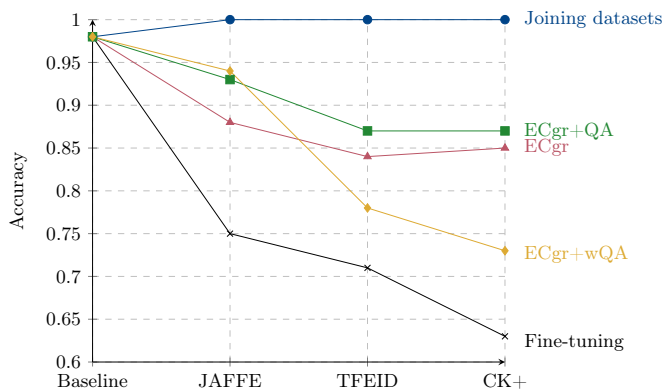

We can better understand the results in the MUG dataset from the continuous training of all datasets with Fig.~\ref{results:accuracy_mug}. It is noticeable that the best method for the MUG dataset is ECgr+QA. We can also observe the poor performance of the fine-tuning method in the context of continuous training, where the knowledge was significantly forgotten compared to methods that attempt to mitigate this behavior. While fine-tuning initially shows promise in adapting the model to new tasks or domains, its performance deteriorates over time as knowledge retention becomes increasingly challenging. Additionally, we can observe that since all datasets are always available, memory forgetting becomes a trivial issue since datasets can be combined to retrain new data. However, one must consider the time and cost of training for such and the storage cost of these datasets.

\begin{figure}[!ht]
    \centering
    \resizebox{0.425\textwidth}{!}{
    \begin{subfigure}{0.45\textwidth}
        \centering
        \begin{tikzpicture}
        \definecolor{ptblue}{RGB}{0,68,136}
        \definecolor{ptred}{RGB}{187,85,102}
        \definecolor{ptyellow}{RGB}{221,170,51}
        \definecolor{ptgreen}{RGB}{34,136,51}
            \begin{axis}[
                clip=false,
                width=\textwidth, % Set the width of the plot to match the subfigure
                ylabel={Accuracy},
                xtick=data,
                xticklabels={Baseline, TFEID, CK+},
                ymin=0.45, ymax=1,
                ytick={0.5,0.6,0.7,0.8,0.9,1},
                axis lines=left,
                xmajorgrids=true,
                ymajorgrids=true,
                grid style=dashed,
                ]
                \addplot[color=ptblue,mark=*] coordinates {
                    (1, 0.7800)
                    (2, 0.9400)
                    (3, 0.9900)
                };
                %\addlegendentry{Joint}
                
                \addplot[color=black,mark=x] coordinates {
                    (1, 0.7800)
                    (2, 0.7600)
                    (3, 0.5700)
                };
                %\addlegendentry{Fine-tune}
                
                \addplot[color=ptred,mark=triangle*] coordinates {
                    (1, 0.7800)
                    (2, 0.6400)
                    (3, 0.6100)
                };
                %\addlegendentry{ECgr}
                
                \addplot[color=ptgreen,mark=square*] coordinates {
                    (1, 0.7800)
                    (2, 0.6200)
                    (3, 0.5500)
                };
                %\addlegendentry{ECgr+QA}
                
                \addplot[color=ptyellow,mark=diamond*] coordinates {
                    (1, 0.7800)
                    (2, 0.6900)
                    (3, 0.5900)
                };
                %\addlegendentry{ECgr+wQA}
            \end{axis}
        \end{tikzpicture}
        \caption{Results for the JAFFE dataset}
        \label{results:accuracy_jaffe}
    \end{subfigure}
    }
    ~
    \resizebox{0.425\textwidth}{!}{
    \begin{subfigure}{0.45\textwidth}
        \centering
        \begin{tikzpicture}
        \definecolor{ptblue}{RGB}{0,68,136}
        \definecolor{ptred}{RGB}{187,85,102}
        \definecolor{ptyellow}{RGB}{221,170,51}
        \definecolor{ptgreen}{RGB}{34,136,51}
            \begin{axis}[
                clip=false,
                width=\textwidth, % Set the width of the plot to match the subfigure
                xtick=data,
                xticklabels={Baseline, CK+},
                ymin=0.45, ymax=1,
                ytick={0.5,0.6,0.7,0.8,0.9,1},
                legend pos=outer north east, % Move the legend to the top left corner
                axis lines=left,
                xmajorgrids=true,
                ymajorgrids=true,
                grid style=dashed,
                mark size=2.5,
                ]
                \addplot[color=ptblue,mark=*] coordinates {
                    (1, 0.7400)
                    (2, 1.0000)
                }node[right,pos=1.03]{Joining};
                %\addlegendentry{Joint}
                
                \addplot[color=black,mark=x] coordinates {
                    (1, 0.7400)
                    (2, 0.4900)
                }node[right,pos=1.03]{Fine-tuning};
                %\addlegendentry{Fine-tune}
                
                \addplot[color=ptred,mark=triangle*] coordinates {
                    (1, 0.7400)
                    (2, 0.6200)
                }node[right,pos=1.03]{ECgr};
                %\addlegendentry{ECgr}
                
                \addplot[color=ptgreen,mark=square*] coordinates {
                    (1, 0.7400)
                    (2, 0.7600)
                }node[right,pos=1.03]{ECgr+QA};
                %\addlegendentry{ECgr+QA}
                
                \addplot[color=ptyellow,mark=diamond*] coordinates {
                    (1, 0.7400)
                    (2, 0.7000)
                }node[right,pos=1.03]{ECgr+wQA};
                %\addlegendentry{ECgr+wQA}
            \end{axis}
        \end{tikzpicture}
        \caption{Results for the TFEID dataset}
        \label{results:accuracy_tfeid}
    \end{subfigure}
    }
    \caption{Accuracy comparison for the JAFFE and TFEID datasets.}
    \label{results:jaffe_tfeid_accuracy}
    \vspace{-15pt}
\end{figure}
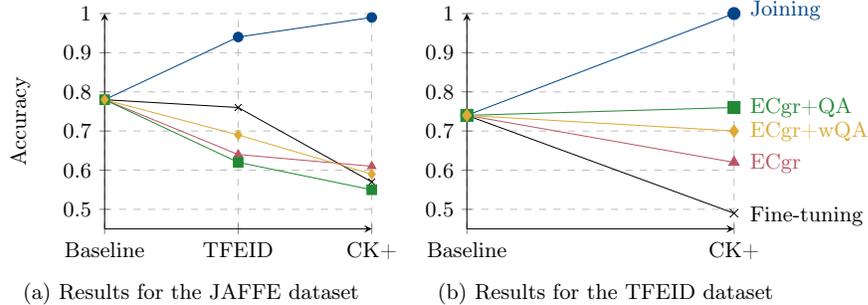

In this same scenario, we can observe the results for JAFFE and TFEID datasets as the original CNN - trained on the MUG dataset - is adapted to the new datasets. In these cases, the baseline is considered the test result obtained from fine-tuning the CNN retrained for the dataset for which the results are reported. Figs.~\ref{results:accuracy_jaffe} and \ref{results:accuracy_tfeid} demonstrate these results. We do not report the results of the CK+ dataset in this scenario because it is the last dataset in the continuous training. Therefore, it does not have subsequent results after its retraining. Please refer to Table \ref{results:table_mugjaffetfeid_to_ckold} for CK+ results.

\vspace{-10pt}

% figura com o gráfico da jaffe e tfeid

\begin{figure}[!ht]
    \centering
    \begin{tikzpicture}[scale=0.8]
    \definecolor{ptblue}{RGB}{0,68,136}
    \definecolor{ptred}{RGB}{187,85,102}
    \definecolor{ptyellow}{RGB}{221,170,51}
    \definecolor{ptgreen}{RGB}{34,136,51}
    \definecolor{ptpurple}{RGB}{170,51,119}
    \definecolor{ptcyan}{RGB}{102,204,238}
        \begin{axis}[
            clip=false,
            ylabel={F1 score},
            ymin=0.4,
            legend columns=3, 
            legend style={at={(0.5,-0.13)},anchor=north},
            xtick=data,
            xticklabels={Baseline, JAFFE, TFEID, CK+},
            axis lines=left,
            xmajorgrids=true,
            ymajorgrids=true,
            grid style=dashed,
            mark size=2,
            ]
            \addplot[dashed,color=red,mark=square*] coordinates {
                (1, 0.9800)
                (2, 0.5200)
                (3, 0.5200)
                (4, 0.4400)
            };%node[right, pos=1.03]{Joining datasets};
            \addlegendentry{Fine-tuning (anger)}
            
            \addplot[dashed,color=ptgreen,mark=square*] coordinates {
                (1, 0.9900)
                (2, 0.8900)
                (3, 0.7400)
                (4, 0.4200)
            };%node[right, pos=1.03]{Joining datasets};
            \addlegendentry{Fine-tuning (disgust)}

            \addplot[dashed,color=ptpurple,mark=square*] coordinates {
                (1, 0.9500)
                (2, 0.8800)
                (3, 0.9200)
                (4, 0.9100)
            };%node[right, pos=1.03]{Joining datasets};
            \addlegendentry{Fine-tuning (fear)}

            \addplot[dashed,color=ptyellow,mark=square*] coordinates {
                (1, 1.0000)
                (2, 0.9400)
                (3, 0.9100)
                (4, 1.0000)
            };%node[right, pos=1.03]{Joining datasets};
            \addlegendentry{Fine-tuning (happiness)}

            \addplot[dashed,color=ptblue,mark=square*] coordinates {
                (1, 0.9800)
                (2, 0.9300)
                (3, 0.8900)
                (4, 0.9100)
            };%node[right, pos=1.03]{Joining datasets};
            \addlegendentry{Fine-tuning (sadness)}

            \addplot[dashed,color=ptcyan,mark=square*] coordinates {
                (1, 0.9900)
                (2, 0.9600)
                (3, 0.9900)
                (4, 0.9800)
            };%node[right, pos=1.03]{Joining datasets};
            \addlegendentry{Fine-tuning (surprise)}

            %\addplot[dashed,color=black,mark=square*] coordinates {
            %    (1, 0.9700)
            %    (2, 0.3100)
            %    (3, 0.2500)
            %    (4, 0.1900)
            %};%node[right, pos=1.03]{Joining datasets};
            %\addlegendentry{Fine-tuning (neutral)}

            \addplot[color=red,mark=*] coordinates {
                (1, 0.9800)
                (2, 0.9200)
                (3, 0.7000)
                (4, 0.4700)
            };%node[right, pos=1.03]{Joining datasets};
            \addlegendentry{ECgr+QA (anger)}
            
            \addplot[color=ptgreen,mark=*] coordinates {
                (1, 0.9900)
                (2, 0.9600)
                (3, 0.9300)
                (4, 0.8600)
            };%node[right, pos=1.03]{Joining datasets};
            \addlegendentry{ECgr+QA (disgust)}

            \addplot[color=ptpurple,mark=*] coordinates {
                (1, 0.9500)
                (2, 0.9100)
                (3, 0.8000)
                (4, 0.8000)
            };%node[right, pos=1.03]{Joining datasets};
            \addlegendentry{ECgr+QA (fear)}

            \addplot[color=ptyellow,mark=*] coordinates {
                (1, 1.0000)
                (2, 0.9700)
                (3, 0.9700)
                (4, 0.9300)
            };%node[right, pos=1.03]{Joining datasets};
            \addlegendentry{ECgr+QA (happiness)}

            \addplot[color=ptblue,mark=*] coordinates {
                (1, 0.9800)
                (2, 0.9400)
                (3, 0.9200)
                (4, 0.9100)
            };%node[right, pos=1.03]{Joining datasets};
            \addlegendentry{ECgr+QA (sadness)}

            \addplot[color=ptcyan,mark=*] coordinates {
                (1, 0.9900)
                (2, 0.9800)
                (3, 0.9600)
                (4, 0.9900)
            };%node[right, pos=1.03]{Joining datasets};
            \addlegendentry{ECgr+QA (surprise)}

            %\addplot[color=black,mark=*] coordinates {
            %    (1, 0.9700)
            %    (2, 0.9400)
            %    (3, 0.8700)
            %    (4, 0.4100)
            %};%node[right, pos=1.03]{Joining datasets};
            %\addlegendentry{ECgr+QA (neutral)}
        \end{axis}
    \end{tikzpicture}
    \caption{Comparison of F1 scores by class on the MUG dataset between fine-tune and ECgr+QA, showcasing the continuous adaptation across JAFFE, TFEID, and CK+ datasets.}
    \label{results:f1_classes_finetune_ecgr+qa}
\end{figure}
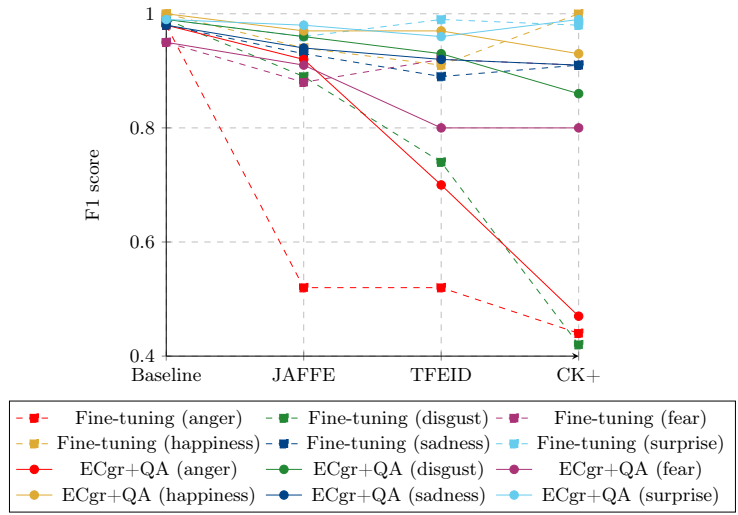

Given that synthetic images for each class are generated independently in our method, it is essential to examine class-specific memory loss. Figure \ref{results:f1_classes_finetune_ecgr+qa} compares the fine-tune and ECgr+QA methods, revealing subpar performance (F1 < 0.6) for the anger and disgust classes. The ECgr+QA method also experiences performance deterioration for the fear class during the final retraining step. This underscores the difficulty associated with training these classes, as even minor facial changes can be misinterpreted as another emotional state.

\vspace{-5pt}

\section{Discussion}\label{sec5}

% TODO
%\Alceu{Seria interessante alguma análise a nível de classes? Digo, apresentar o impacto ao longo do processo (perdas) por classe? Teria como criar uma figura mostrando a matriz de confusão antes e no final de todas as iterações?}

Firstly, the results support our hypothesis regarding using pseudo-rehearsal methods, specifically emotion-centered generative replay, to minimize memory decay. Our strategy demonstrated remarkable efficacy in alleviating catastrophic forgetting, consistently outperforming the fine-tuning methods across various tasks. The generation of synthetic data resembling past task patterns through WGAN-GPs proved positive in enabling the network to retain knowledge without using original data. This substantiates our anticipation that pseudo-rehearsal techniques, particularly our emotion-centered generative replay, play an important role in counteracting memory decay.

Furthermore, synthesizing our WGAN-GP class-driven generative and QA methods substantiates our second hypothesis. Introducing a QA mechanism during replay significantly improved the quality of synthetic data, further augmenting the approach's effectiveness. The third hypothesis, in which applying a weight to synthetic images would benefit continuous training, can only be observed as positive in the retraining for the first dataset - from MUG to JAFFE. We observed that this technique was ineffective for more datasets subsequent to JAFFE. This may be directly related to the errors of the network that assigns these weights to the synthetic images - that is, the network may be making errors with high confidence, and this negatively affects the synthetic images, which in turn are not fully considered in the retraining, leading the CNN to not `remember' these data.

\vspace{-30pt}

\section{Conclusion}\label{sec6}

In this study, we presented a comprehensive investigation into the challenge of catastrophic forgetting in CNNs within the context of facial expression recognition, proposing a novel approach to mitigate its effects. We employed a pseudo-rehearsal method, specifically our emotion-centered generative replay (ECgr) with WGAN-GPs, to generate synthetic images for each class of the datasets and combined this with a filtering method to exclude images that could hinder retraining.

Across various tasks, ECgr consistently demonstrated superior performance compared to baseline and fine-tuned methods. Utilizing WGAN-GPs to synthesize task-specific data, along with our QA algorithm, resulted in substantial knowledge retention. This confirms the potential of pseudo-rehearsal methods to effectively retrain CNNs without revisiting original datasets, offering a promising strategy for addressing memory decay, particularly in challenging scenarios like facial expression recognition.

%However, our investigation also revealed certain limitations. While our approach showed promising results, it is important to acknowledge that the effectiveness of pseudo-rehearsal may vary depending on specific network architectures, datasets, and task complexities. Additionally, the computational requirements of WGAN-GP-based - or any GAN-based - data generation may present challenges in real-time applications with strict computational constraints. These aspects suggest potential areas for future enhancements and research directions. Moreover, considering the results obtained using weights for regularizing synthetic images, we may explore better algorithms for assigning these weights during training. Also, investigating the interplay of emotion-centered generative replay with regularization techniques could yield valuable insights.

%In conclusion, our study represents significant progress in addressing the challenge of catastrophic forgetting in neural networks within the context of facial emotion recognition. This work contributes to the expanding body of research to develop more adaptable and memory-retentive neural network models, offering promise for their application in dynamic and evolving environments. As the field of neural network research continues to grow, our findings underscore the importance of integrating innovative strategies to overcome challenges that impede their practical deployment.

Despite promising results with pseudo-rehearsal, its effectiveness may vary across network architectures, datasets, and tasks. Additionally, WGAN-GP-based data generation can be computationally expensive, limiting real-time use. These aspects highlight opportunities for future research, such as improved weight assignment algorithms and exploration of regularization techniques' synergy with pseudo-rehearsal approaches.

%\subsubsection{Acknowledgements} Please place your acknowledgments at the end of the paper, preceded by an unnumbered run-in heading (i.e. 3rd-level heading).

\bibliographystyle{splncs04}
\bibliography{references/references}

\end{document}